# A maintenance planning framework using online and offline deep reinforcement learning

Zaharah A. Bukhsh[1] · Hajo Molegraaf[2] · Nils Jansen[3]

## Abstract

Cost-effective asset management is an area of interest across several industries. Specifically, this paper develops a deep reinforcement learning (DRL) solution to automatically determine an optimal rehabilitation policy for continuously deteriorating water pipes. We approach the problem of rehabilitation planning in an online and offline DRL setting. In online DRL, the agent interacts with a simulated environment of multiple pipes with distinct lengths, materials, and failure rate characteristics. We train the agent using deep Q-learning (DQN) to learn an optimal policy with minimal average costs and reduced failure probability. In offline learning, the agent uses static data, e.g., DQN replay data, to learn an optimal policy via a conservative Q-learning algorithm without further interactions with the environment. We demonstrate that DRL-based policies improve over standard preventive, corrective, and greedy planning alternatives. Additionally, learning from the fixed DQN replay dataset in an offline setting further improves the performance. The results warrant that the existing deterioration profiles of water pipes consisting of large and diverse states and action trajectories provide a valuable avenue to learn rehabilitation policies in the offline setting, which can be further fine-tuned using the simulator.

**Keywords** Deep reinforcement learning · Maintenance planning · Water distribution systems · Conservative Q-learning · Deep Q-networks · Offline DRL

## 1 Introduction

Reliable water distribution systems (WDS) are paramount for functioning societies. Such systems are subject to deterioration around the globe due to budget cuts, lack of maintenance, and an increase in urbanization [1]. Different kinds of *maintenance approaches*, such as recurring schedules (planned) or run-to-failure (corrective), are implemented to keep these assets from failing. However, such approaches cannot facilitate effective solutions with a minimal cost at the desired level of services. Replacing an asset according to a predefined plan results in the loss of its useful functioning life, whereas replacement after failure can cause large consequential damage resulting in unavailability of services [2].

The recent success of deep neural networks as high-capacity function approximators, such as deep Q-networks, has stimulated enormous progress in solving sequential decision-making problems. In particular, the integration of deep learning and reinforcement learning as *deep reinforcement learning* (DRL) has been applied to various applications, including board games [3], video games [4], robotic control [5], and optimal routing [6]. DRL harnesses powerful general-purpose representations to learn and characterize feedback in a long-term horizon [7]. Yet, beyond games and standard optimization problems such as knapsack or traveling salesman, application-oriented studies of DRL for practical, real-world problems remain scarce.

✉ Zaharah A. Bukhsh
z.bukhsh@tue.nl

Hajo Molegraaf
hajo.molegraaf@rolsch.nl

Nils Jansen
n.jansen@science.ru.nl

1 Eindhoven University of Technology, Eindhoven, The Netherlands

2 Rolsch Assetmanagement, Enschede, The Netherlands

3 Radboud University, Nijmegen, The Netherlands

This paper develops and implements a DRL solution to automatically devise an optimal rehabilitation policy for WDS under economic and performance requirements. Note that the *policy* in reinforcement learning is the agent's way of learning and behaving at a given time [7]. In this application setting, the learned policy is referred to as *rehabilitation/maintenance plan*. We study the problem of rehabilitation planning via online and offline DRL approaches, illustrated in Fig. 1. In the online learning paradigm, the agent actively interacts with the (simulated) environment to collect experiences to learn an optimal policy. The offline learning paradigm seeks to learn optimal policies using a logged dataset. The need for active interactions and inability to (re)use the large, diverse dataset remains one of the main limitations for the wide applicability of reinforcement learning framework in practical applications [8]. In this paper, we further study whether the optimal policy found by online DRL can be improved further by reusing the logged dataset in the offline DRL setting. Given the deterioration profiles (states) and subsequent maintenance (actions) on assets, offline DRL offers a useful paradigm to explore and learn rehabilitation policies using the logged dataset.

In the following, we explain the online learning setting followed by the offline DRL approach. Figure 1a shows the schematic illustration of the online DRL setup. We model agent and environment interactions as a Markov decision process (MDP) [9]. Note that we do not explicitly create such an MDP within our framework but merely use it to gain insights into the problem. The agent observes the (simulated) environment state $s$ at time $t$ and performs an action $a_t$. The agent receives a reward $r_t$ as a feedback signal, and the environment moves to the next state represented as $s_{t+1}$. The MDP formulation assumes that the state transition follows the Markovian property, meaning that future states depend only on the current state $s_t$ and action $a_t$ irrespective of the agent's previous states and actions. The agent's interactions with the environment, based on a policy, i.e. $\pi_k$, are collected in a buffer denoted as $\mathcal{D}$. The buffer $\mathcal{D}$ consists of samples from multiple policies as $\pi_1, \pi_2, \pi_3, ...\pi_k$, which are further used to update a new policy as $\pi_{k+1}$. We choose a deep Q-network as an online learning algorithm [4].

In an offline setting (see Fig. 1b), the agent does not iteratively collect more data to update the policy. Instead, an offline DRL agent utilizes a static dataset $\mathcal{D}$ consisting of state, action, reward, and next state tuple. The static dataset is generated using a behavior policy depicted as $\pi_\beta$, which is based on either a random policy, an expert policy, or a partially trained online policy. In this paper, we use a dataset accumulated during the training of an online DQN agent based on a near-expert policy. Previous studies [10, 11] also used the dataset collected by an online agent for learning in an offline DRL setup. We use a conservative Q-learning (CQL) algorithm proposed by Kumar et al. [12] to learn an optimal policy from a static dataset.

In both online and offline DRL, the agent seeks to maximize the expected cumulative reward in a definite time horizon following a sequence of actions. Our work offers the following key results and contributions:

– We establish a novel solution for optimal rehabilitation planning of water distribution systems under economic and performance requirements. We design the elements for a DRL framework, including state definition, discrete actions, and reward function.
– We are the first to introduce offline reinforcement learning to solve the practical problem of rehabilitation planning of water pipes using a static dataset only. We adapt the conservative Q-learning algorithm [12] to use logged interactions from the near-expert policy learned by the online DRL agent.
– We show that the online DRL setup can learn a cost-effective intervention policy compared to the traditional preventive, corrective, and greedy schedules.

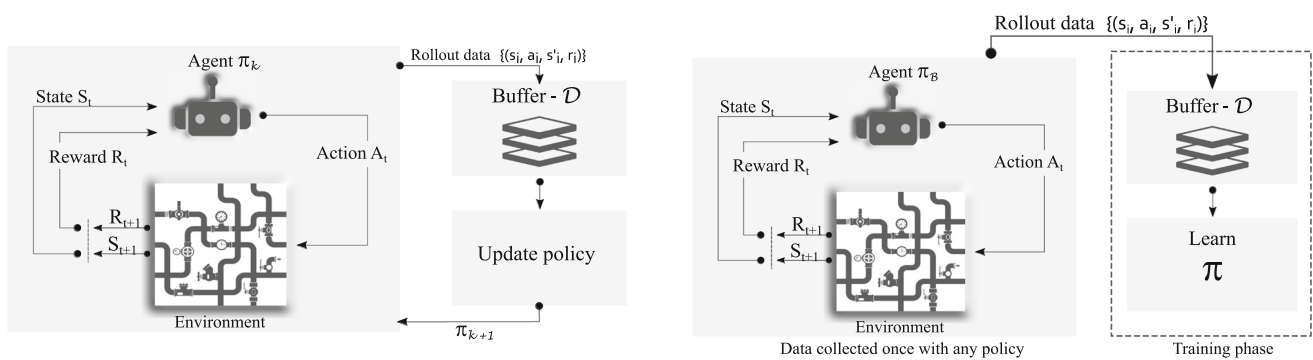

(a) Online (off-policy) deep reinforcement learning.

(b) Offline deep reinforcement learning.

**Fig. 1** Illustration of deep reinforcement learning framework adapted from [8]

Additionally, the offline DRL-based approach can further improve the learned policy using the logged dataset.

The rest of the paper is structured as follows: Sect. 2 provides an overview of related work. Section 3 introduces the problem setting followed by an explanation of solution approaches in Sect. 4. Section 5 introduces the case study, presents the experimental setup, outlines the results, and provides a general reflection, followed by the limitations of the proposed approach. Section 6 highlights the concluding remarks and provides a future outlook.

## 2 Related work

### 2.1 Rehabilitation planning of water pipes

Optimal scheduling for rehabilitation of WDS has been actively studied since 1979 with the pioneering work of Shamir and Howard [13]. The reported methods in the literature employ various optimization techniques and mathematical models such as genetic algorithm [14, 15], dynamic programming [16], integer linear programming [17], multi-criteria decision analysis [18] and budget allocation approaches [19, 20] to facilitate decision-makers in the planning of repair and replacement of water pipes. Advanced technologies such as artificial neural networks (ANN) [21], cluster analysis [22] and graph theory [23] are also adapted by a few studies to identify critical components of WDS and to predict the number of pipe failures along with influencing factors. Similarly, in the realm of sequential decision-making, Markov decision processes (MDP) are used to model the deterioration of water networks [24].

Despite the vast research interests, it is noted that existing planning methods are site-specific and do not include comprehensive criteria (such as economic, reliability, availability, and social impact) of large-scale pipe networks [25, 26]. Besides, traditional programming approaches such as integer programming and evolutionary methods do not scale to accommodate a large number of continuously changing the physical state of pipes, and the (computational) complexity of these methods grows exponentially with the problem size [27].

### 2.2 Deep reinforcement learning applications

With the success of the Deep Q network for playing Atari video games [4], the field of deep reinforcement learning (DRL) has gained enormous traction. DRL has outperformed human experts on several board [3], card [28] and video games [29]. It has also been applied to solve complex robotic movements [5], vision control [30], and routing tasks [6]. Nevertheless, many opportunities (and challenges) of DRL in solving problems in diverse domains such as asset management, health, manufacturing, and transportation have remained under-explored.

A few notable studies address sequential decision-making problems from diverse application domains using the DRL. Zheng et al. [31] conducted a comprehensive examination to learn dynamic tax policies by economic simulations to balance equality and productivity in socio-economic settings. Zheng et al. [32] solved a chemical production scheduling process to account for uncertainty using the actor-critic policy gradient algorithm. Cals et al. [33] introduced an approach to solving the order batching and sequencing problems in a warehouse using the proximal policy optimization algorithm. Similarly, using a multi-agent framework, the sequencing problem to avoid bus bunching is addressed by Wang and Sun [34]. The DRL-based methods are also being employed for the efficient energy management of the buildings [35, 36].

Wei et al. [37] suggest planning of intervention at the asset level with a deterministic environment setting. Lei [38] introduces a life-cycle maintenance planning approach for network-wide bridges. Huang et al. [39] propose a DRL approach for preventive maintenance of production lines. Khorasgani et al. [40] provides an offline reinforcement learning approach for maintenance decision-making. Our work distinguishes itself from previous studies as we investigate the DRL application for the maintenance planning of water pipes under stochastic settings. We aim to minimize the average intervention cost and the failure probability while considering the chances of unplanned failures, whereas previous studies mainly focus on improving assets' performance. Our paper also studies the offline RL paradigm to further improve the policy learned by online variant using the logged dataset.

## 3 Problem formulation

The objective is to learn an optimal rehabilitation policy for continuously deteriorating water pipes. A policy is optimal if it incurs minimal average cost and reduced failure probability for a definite planning horizon. We achieve this by finding optimal intervention moments such that maintenance is performed before failure but not too early, resulting in a waste of functional life and not too late causing system unavailability and additional costs. In the following, we formulate the maintenance planning with an MDP.

*States* The state space represents the (physical) characteristics of a pipe. It consists of a pipe's age, material, failure rate, and failure probability, denoted as $s_t = \langle \text{age, mat}, \lambda, pf \rangle$. The failure rate concerning the material is obtained from [41], whereas the probability of failure is elicited from Eq. 4. Besides material, the failure rate is also dependent on the length of the pipe as longer pipes are likely to experience more failures compared to the smaller pipes [42]. Therefore, the failure rate of each pipe is multiplied by its length to obtain the failure rate per meter of the pipe.

*Actions* The agent's objective is to find an optimal action depending on the given state at each timestep. The action $a_t \in [0, 1, 2]$ represents the discrete actions for an agent at each timestep for a pipe, where $a_t = 0$ suggests the *do nothing* action, $a_t = 1$ represents the *maintain* action, and $a_t = 2$ denotes the *replace* action.

*Reward function* The DRL-based agent seeks to maximize the expected cumulative reward in a definite time horizon following a sequence of actions [7]. Accordingly, we design the reward function to represent our user-specific objective. We construct the reward function with inverse values that seek to minimize the overall intervention cost and failure probability, as shown in the following Equation.

$$R(s_t, a_t, s_{t+1}) = \text{MC}_t + (-pf_t) \quad (1)$$

where $\text{MC}_t$ represents the maintenance intervention cost, and $pf_t$ is the failure probability, corresponding to the safety risk for the asset at each timestep $t$. The cost is computed as follows:

$$\text{MC}_t = \begin{cases} 0 & \text{if } a_t = 0 \\ -0.5 & \text{if } a_t = 1 \text{ and } pf_t > 0.5 \\ -0.8 & \text{if } a_t = 2 \text{ and } pf_t > 0.5 \\ -1 & \text{if } a_t = 0 \text{ and } pf_t > 0.9 \\ -1 & \text{if } a_t = 1 \text{ and } pf_t \leq 0.5 \\ -1 & \text{if } a_t = 2 \text{ and } pf_t \leq 0.5 \end{cases} \quad (2)$$

We introduce a penalty of $-1$ to discourage unnecessary maintenance actions. Similarly, a penalty is given if the system is near failure, yet the agent chooses the action of *do nothing*. Note that the cost values in the reward function are representative and do not relate to the material and length of the pipe.

*Dynamics (Time to transition)* The environment simulates the physical characteristics of water pipes. At any timestep, the agent receives a representation of the environment's state in the form of state $s_t \in \mathcal{S}$ where $s_t = \langle \text{age, mat}, \lambda, pf \rangle$. The agent responds with an action $a_t \in \mathcal{A}$, receives a reward $r_t \in \mathcal{R}$ and moves to next state $s_{t+1} \in \mathcal{S}$. In the finite MDP, the next state $s_{t+1}$ and reward $r_{t+1}$ have a discrete probability distribution, dependent only on the current $s_t$ and $a_t$ [7]. The dynamics function (also referred to as state transition probability function) is defined as:

$$p(s', r|s, a) = \mathbb{P}(s_{t+1} = s', r_{t+1} = r|s_t = s, a_t = a) \quad (3)$$

In other words, the next state and reward are only dependent on the current state and action irrespective of all the previously visited states, thus respecting the Markov property.

We estimate the failure probability of water pipes using the exponential (Poisson) distribution [43] represented as:

$$pf_t = 1 - e^{-\lambda \times \text{age}_t} \quad (4)$$

where $pf_t$ is the probability of failure at time $t$, and $\lambda$ is the failure rate, and $\text{age}_t$ is the current age of the pipe.

In each iteration, the $\text{age}_t$ of the pipe is updated depending on the agent's chosen action as follows:

$$\text{age}_{t+1} = \begin{cases} \text{age}_t + 1 & \text{if } a_t = 0 \\ \text{age}_t - U(j, k) & \text{if } a_t = 1 \\ \text{age}_t = 1 & \text{if } a_i = 2 \end{cases} \quad (5)$$

The age variable $\text{age}_{t+1}$ is

- incremented with one (year) in case of *doing nothing*, depicting the increase in *pf*.
- reduced by a certain factor sampled from a uniform distribution between $j$ and $k$ as a result of the *maintain* action. This is to simulate the variable improvement in the pipe's physical state, which depends on several factors such as its material, length, intervention type, and intervention quality [38].
- set to one to depict the good as the new condition state of the pipe resulting from the *replace* action.

Besides a gradual degradation, an asset can experience sudden failure due to changes in its surroundings and environmental impacts. Therefore, in each iteration, we simulate a 5% random chance of sudden failure leading to the obvious replacement action.

# 4 Reinforcement learning approaches

This section briefly explains deep Q-network (DQN) and conservative Q-learning (CQL) algorithms used to solve the maintenance planning problem. CQL is an offline DRL algorithm comparable to DQN in an online DRL setting.

## 4.1 Deep Q-network for online DRL

The agent's interactions with the environment generate a sequence of trajectories, which are denoted as:

$$s_0, a_0, r_1, s_1, a_1, r_2, s_2, a_2, r_3, \ldots \quad (6)$$

The goal of an agent is to find an optimal policy that maximizes the expected accumulated return $G_t$, which is discounted sum of rewards represented as $G_t = r_t + \gamma r_{t+1} + \gamma^2 r_{t+2}...$, where $r_t$ is the reward received at time $t$, and $\gamma \in [0, 1]$ is a discount factor. The optimal Q-function (action-value function) must provide the maximum action values at all the states determined by the Bellman optimality Equation as follows [7]:

$$Q^*(s, a) = \mathbb{E}_{s'}[r + \gamma \max_{a'} Q^*(s', a')] \qquad (7)$$

where $r$ is the immediate reward received, $a'$ is the action to move to a state $s'$ that returns the maximum reward. $\mathbb{E}$ is the expected value of a random variable given that the agent follows the policy $\pi$. The optimal policy $\pi$ is derived by Eq. 7 in an iterative update such that the agent starts in state $s$ and takes the highest return action $a$ and follows the policy $\pi$ for all future steps.

Q-learning falls short for complex systems beyond standard grid examples. The seminal work of [4] proposed to utilize deep neural networks (DQN) as function approximators for estimating Q-functions for high-dimensional state spaces. For a long time, utilizing neural networks for reinforcement learning remained an open research question due to the instability of network training and convergence caused by correlation among observations [44]. DQN introduced two training techniques. First, it implements *experience replay* that randomly samples observations, thus avoiding correlations. Second, the target value $[r + \gamma \max_{a'} Q(s', a')]$ is only periodically updated against the action values $Q(s, a)$.

The value function, which uses neural networks as approximator, is denoted as $Q(s, a; \theta_i)$ where $\theta$ are weights of Q network at i$^{th}$ iteration. The Q-value is updated using the following loss function [4]:

$$L_i(\theta_i) = \mathbb{E}_{(s,a,r,s') \sim U(D)} \\ \left[ \left(r + \gamma \max_{a'} Q(s', a'; \theta_i^-) - Q(s, a; \theta_i)\right)^2 \right] \qquad (8)$$

where $U(D)$ represents the uniform distribution over the transitions tuple $(s, a, r, s')$ drawn from experience replay to apply the Q-value updates. The $\gamma$ is the discount factor, $\theta_i$ are the weights (the neural network) used to determine the $Q(s, a|\theta)$ and $\theta_i^-$ are the weights to compute the target value. As noted earlier, the $\theta_i^-$ weights are updated only periodically.

## 4.2 Conservative Q-learning for offline RL

Offline DRL aims to learn optimal policy $\pi$ that returns maximum discounted future reward by learning only from a static dataset having transitions like $(s, a, r, s')$. The static dataset is generated using behavior policy $\beta$ (as shown in Fig. 1b), which is either based on a random initial policy, near-expert online policy, or their combination [8].

Standard Q-learning methods based on temporal difference suffer from distributional shifts as the learned policy $\pi_k$ differs from the policy $\pi_\beta$ used to collect data [45]. Specifically, when evaluating the next state $s'$ to obtain the action with the highest return, we may be querying the dataset for the $(s', a')$ pairs for which the true Q-value does not exist. Typical off-policy algorithms tend to overestimate the Q-values for unseen actions, thus resulting in the problem of out-of-distribution actions. Distributional shifts, along with sampling and function approximator errors, lead to overestimating the value function in the offline DRL setting. Kumar et al. [12] proposed an effective algorithm called conservative Q-learning (CQL) to address the problem of Q-value overestimation for unseen actions. CQL suggests being conservative (i.e., assigning lower values) in estimating the Q-values and learning the lower bound on its true values. The lower-bounded Q-values guarantee that the learning policy prevents the execution of unseen actions or avoids their overestimation.

CQL estimates the value function using the given dataset $\mathcal{D}$. To avoid overestimation of Q-values, a penalty is introduced to minimize the expected Q-value under a particular state-action pair distribution denoted a $\mu(s, a)$. The authors [12] prove that this minimization value along with the Bellman error objective achieves a lower-bound $Q^\pi$ at all $S, \mathcal{D}, a \in \mathcal{A}$. An additional Q-value maximization term is introduced under the data distribution, $\pi_\beta(a|s)$ for a tighter lower bound. A trade-off factor $\alpha \geq 0$ accounts for lower bounds for true and expected Q-function under sampling error and function approximation. With the availability of a larger dataset $\mathcal{D}$, the lower bounds can be achieved with a minimal $\alpha$ value. The mathematical formulation for conservative policy evaluation is given below:

$$\hat{Q}_{CQL}^\pi \leftarrow \underset{Q}{\arg \min} \, \alpha \\ \left( \mathbb{E}_{s \sim \mathcal{D}, a \sim \mu(a|s)}[Q(s, a)] - \mathbb{E}_{s \sim \mathcal{D}, a \sim \hat{\pi}_\beta(a|s)}[Q(s, a)] \right) \\ + \frac{1}{2} \mathbb{E}_{(s,as')\sim\mathcal{D}} \left[ \left( Q(s, a) - \mathcal{B}^{\hat{\pi}} Q \right)^2 \right] \qquad (9) \\ \hat{\mathcal{B}}^\pi Q = r + \gamma \mathbb{E}_\pi[Q(s', a')].$$

where $\mathcal{D} = (s, a, r, s')$ is a dataset of tuples from trajectories collected using the behavior policy $\pi_\beta(a|s)$. Since $\mathcal{D}$ does not contain all the state action transitions, the policy evaluation uses an empirical bellman operator denoted as $\mathcal{B}^{\hat{\pi}}$ and the $\hat{\pi}_\beta(a|s)$ denotes the empirical behavior of the policy. The learned policy using $\hat{Q}_{CQL}^\pi$ can be used for policy optimization in a typical temporal difference learning procedure.

# 5 Case study and results

We use the data of 16 pipes from the WDS, given in Table 1. Each pipe has an age, material, length, failure rate, and failure probability computed using Eq. 4. The failure rate is determined based on pipe material. The length information is used to calculate the failure rate per meter of the pipe as the larger pipe is likely to experience a higher number of failures compared to smaller pipes [42]. During training, we randomly sample a single pipe detail for learning in each episode. This is to ensure that the agent is introduced with diverse initial states. Once the agent is trained, we evaluate its performance on all the pipes and report the average cost and failure probabilities.

## 5.1 Experiment setup for DQN

We develop a simulation environment of water pipes using the standard Open AI Gym library. We train an agent to converge to an optimal rehabilitation policy of 100 years, where each timestep is a year. An episode finishes when the timestep reaches the $100^{th}$ year. We evaluate the quality of actions in a single trajectory (episode) by computing the discounted sum of rewards $G_t$. We use standard DQN algorithm implementation from StableBaseline library [46].

We performed hyperparameter tuning to find optimal parameters for training the agent. We assess the impact of a single parameter on the performance by altering its value while the remaining parameters remains fixed. This approach enables us to study the impact of a single parameter on the performance of the agent. Table 2 shows the base parameters configuration, tested parameters ranges, and the chosen value for the DQN agent. We study the significance of network architecture, activation function, buffer size, discount factor, the minimum value for epsilon, and learning rate. Figure 2 shows the rolling mean and standard deviation of return obtained by the agent for 1000 training episodes under base parametric settings noted in Table 2. To study the impact of a larger buffer size, the agent is trained longer with 10000 episodes. The performance remains comparable for most parameters except for the *Tanh* activation function, and extremely small learning rates negatively impact the performance. The larger buffer size also does not result in improved performance. We noted that the few agents, e.g., one trained with (100, 50, 25) network configuration performed well in the training environment; however, it performed poorly in the test environment due to changes in values of parameters resulting from hyperparameter tuning. Therefore, the best-performing parameters are not always chosen for training the agent.

The training performance of DQN agent with chosen parameters is shown in Fig. 3. Note that the cost represents the intervention and penalty for higher failure probability incurred for a horizon of 100 timesteps for a single pipe. As can be noticed, the performance improves after 600 episodes and stabilizes after around 750 episodes.

## 5.2 Experiment setup for CQL (Offline)

Given that each pipe can have three discrete actions, we implement discrete CQL with DQN using the offline reinforcement learning library [47]. The offline learning agent uses a static dataset $\mathcal{D}$ to optimize an objective

**Table 1** Case study data of water pipes network used for DRL agent training

| Pipes | Age (year) | Material | Length (m) | Failure rate (km) | Failure probability |
|-------|------------|----------|------------|-------------------|---------------------|
| 1 | 44 | Asbestos cement | 2365 | 0.06 | 0.998 |
| 2 | 46 | Asbestos cement | 2732 | 0.06 | 0.999 |
| 3 | 6 | Asbestos cement | 1908 | 0.06 | 0.497 |
| 4 | 42 | Asbestos cement | 1996 | 0.06 | 0.993 |
| 5 | 32 | Ductile iron | 1968 | 0.02 | 0.997 |
| 6 | 37 | Ductile iron | 2915 | 0.02 | 0.884 |
| 7 | 25 | Ductile iron | 2405 | 0.02 | 0.700 |
| 8 | 47 | Ductile iron | 1500 | 0.02 | 0.654 |
| 9 | 11 | Gray cast iron | 2017 | 0.07 | 0.788 |
| 10 | 30 | Gray cast iron | 1679 | 0.07 | 0.971 |
| 11 | 31 | Gray cast iron | 2071 | 0.07 | 0.989 |
| 12 | 45 | Gray cast iron | 1666 | 0.07 | 0.995 |
| 13 | 15 | PVC | 1650 | 0.015 | 0.310 |
| 14 | 40 | PVC | 2365 | 0.015 | 0.758 |
| 15 | 22 | PVC | 2434 | 0.015 | 0.552 |
| 16 | 2 | PVC | 1527 | 0.015 | 0.045 |

**Table 2** Hyperparameter values for training online DQN agent

| Parameter | Base | Range | Chosen |
|---|---|---|---|
| NN | [16, 32] | [16, 32], [64, 64], [100, 50, 25] | [64, 64] |
| Activation func | ReLU | ReLU, Tanh, LeakyReLU | ReLU |
| Buffer size | $5 \times 10^4$ | $10^4, 5 \times 10^4, 10^5, 10^6$ | $5 \times 10^4$ |
| Discount factor | 0.99 | 0.90, 0.99, 0.999 | 0.99 |
| Clipping epsilon | 0.05 | 0.1, 0.05, 0.025 | 0.1 |
| Learning rate | $10^{-4}$ | $10^{-3}, 10^{-4}, 10^{-5}$ | $10^{-4}$ |

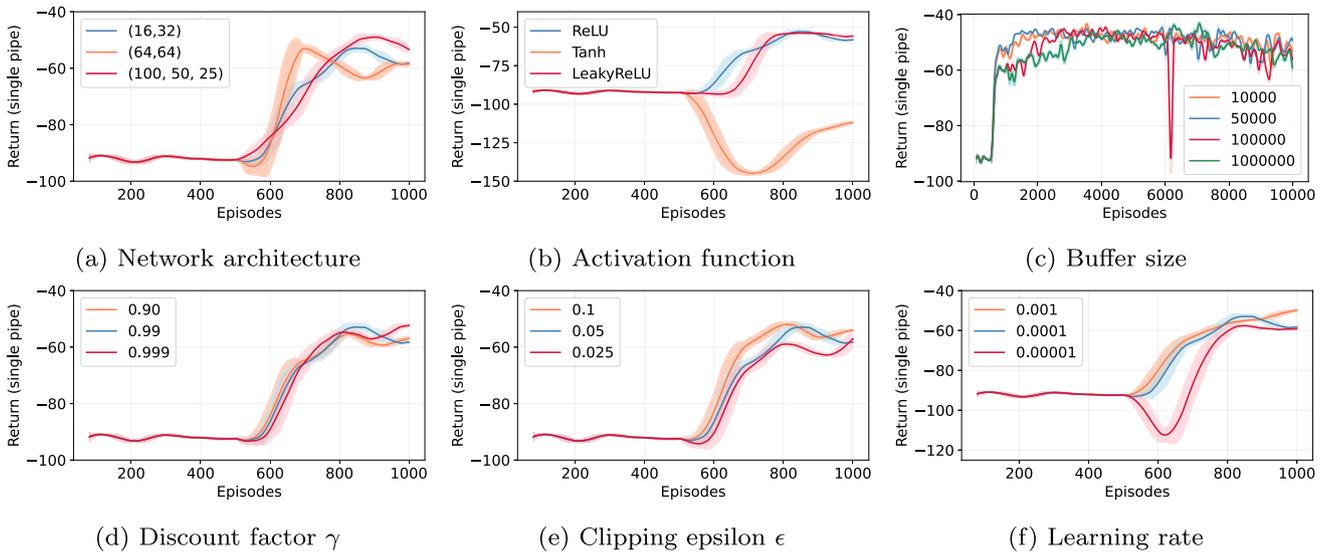

(a) Network architecture   (b) Activation function   (c) Buffer size

(d) Discount factor $\gamma$   (e) Clipping epsilon $\epsilon$   (f) Learning rate

**Fig. 2** Significance of various parameters on the performance of online agent: Each point is a rolling mean, and the shaded area is a rolling standard deviation return with the sliding window of 20 episodes

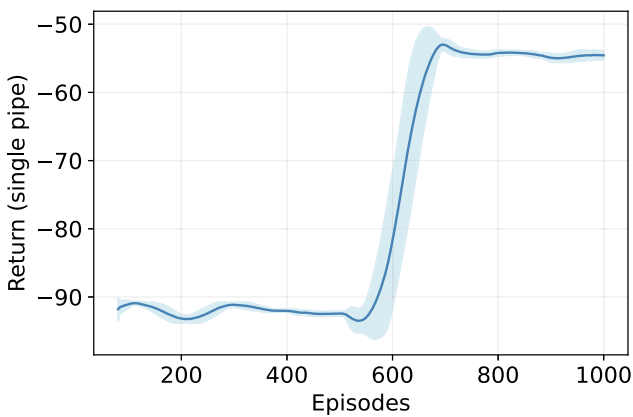

**Fig. 3** The DQN agent performance during training with $\epsilon$-greedy policy. The graph reports the mean and standard deviation of cumulative reward (return) with a sliding window of 20 episodes

function without additional interactions from the environment.

We use a dataset accumulated during the training of an online DQN agent, as shown in previous works [10, 11].

The dataset collected during the training is based on the near-expert policy. The dataset consists of 1000 episodes, each with 100 transitions. A single transition is a tuple of state, action, reward, and next state denoted as $(s_t^i, a_t^i, r_t^i, s_{t+1}^i)$. The dataset is collected once and is not altered during the training and inference. In an offline setting, the episode concept is similar to the online DRL variant, which consists of all the data collected by the agent's active interactions with the environment until a terminating condition is met. Recall that in our case, the agent interacts with the environment for 100 timesteps, where each timestep represents a year. Similar to supervised learning, we split 80% of the dataset for training, and the rest 20% is used for the evaluation after each epoch. We train an offline learning agent to compute the optimal rehabilitation policy for each pipe for 200 epochs. The test dataset is used to evaluate the learning progress of the CQL agent by computing the return.

We study the impact of various network architectures, activation functions, discount factor values, Q-functions, and learning rates. The base parameters, range, and chosen value are noted in Table 3. Figure 4 provides the rolling mean and standard deviation of return obtained by the agent for 200 training epochs under given parametric settings for a single pipe. We notice a higher standard deviation for almost all chosen parameter values than the online setup. We found that most of the parameters perform comparably after 200 epochs of training except for the *Tanh* function.

Figure 5 shows the training curve of the CQL agent. The objective of offline agents is to develop an understanding of the underlying MDP of $\mathcal{D}$ to construct optimal policy entirely from the static dataset. Since the dataset is based on near-optimal policy, the learning agent converges after fewer training epochs. The agent is trained using the best parametric configurations provided in Table 3.

We also compare the CQL learning performance with different source datasets where random policy represents the data collected by the agent's random interaction; the near-expert policy collects the data during training of DQN in which the agent actively explores and exploits the learned information. The expert policy collects the dataset using the trained DQN agent. The training curves are shown in Fig. 6. We found that offline learning using the expert policy dataset results in poor performance. This is because the expert policy dataset can be limited in assessed actions, limiting the agent's ability to experience various actions and resulting rewards. Learning from the dataset from near-expert policy shows good performance in fewer training epochs. It is also noted that the random policy can achieve comparable results to the near-expert policy if trained for longer. This result is aligned with the findings reported in [12, 48].

## 5.3 Results and analysis

We evaluate both online and offline learning reinforcement learning paradigms for the case of maintenance planning. We report the average intervention cost and failure probability obtained for all 16 pipes using the trained DRL agents and baseline policies. To enable comparison with baselines, the intervention cost does not include the penalty of delaying the maintenance action as defined in the reward function. Both trained agents prefer the `replace` action over the `maintain` action. This could be because `replace` action brings an immediate reduction in failure probability compared to `maintain` action. The CQL agent suggests the `replace` action 89 times with approximately 5.56 actions per pipe. The DQN agent takes `replace` actions 122 times with a mean value of 7.6 per pipe. Since `do nothing` action incurs no cost, this action is predominant in both rehabilitation plans, which is also aligned with the real planning situation.

We establish baselines with time-based preventive, corrective, and greedy approaches to evaluate and compare the usefulness of employing the DRL framework for rehabilitation planning. The time-based preventive planning approach is based on a recurring schedule, whereas in the corrective approach, mainly `replace` action is executed after the failure. The corrective approach results in higher costs due to the impact on the network and users. However, in our comparison, we only include the cost of replacement. The greedy approach takes locally optimal choices based on heuristics. We develop a rehabilitation plan for all pipes for 100 per year with a time-based preventive approach, where `maintain` action is performed every five (referred to as Maintain-5) and ten years (referred to as Maintain-10). The action `replace` is performed in the corrective approach when a pipe has $pf_i \geq 0.95$. In the greedy approach, the `maintain` as the cheapest intervention option is chosen as soon as a pipe reaches to $pf_i \geq 0.80$. For a fair comparison with the DRL-based approach, we also consider a random chance of sudden failure leading to replacement action. Additionally, the penalty cost is deducted from DRL methods, and only the intervention costs are reported to enable comparison with the baselines.

Figure 7 shows the average costs and failure probability obtained by developing rehabilitation policy using different methods. We report the average intervention cost and failure probability of all 16 pipes. The DQN approach results in the lowest failure probability. The highest failure probabilities are noted with time-based preventive schedules, which are standard in the practice. This possible justification for higher failure probabilities is that *maintain* action does not bring standard improvement to the performance state of an asset, like a replacement. Instead, it depends on the quality of maintenance actions (see Eq. 5). Besides reducing failure probability, a suitable rehabilitation plan must also incur a minimal cost. The CQL method finds the most cost-effective plan followed by DRL-DQN and Maintain-10. The greedy method provides the most expensive plan with a failure probability of 0.84.

| | | | | |
|---|---|---|---|---|
| **Table 3** Hyperparameter values for training offline CQL agent | Parameter | Base | Range | Chosen |
| | NN | [16, 32] | [16, 32], [64, 64], [100, 50, 25] | [64, 64] |
| | Activation func | ReLU | ReLU, Tanh, LeakyReLU | ReLU |
| | Discount factor | 0.99 | 0.90, 0.99, 0.999 | 0.99 |
| | Q function | Mean | Mean, QR, IQR | Mean |
| | Learning rate | $10^{-4}$ | $10^{-3}$, $10^{-4}$, $10^{-5}$ | $10^{-4}$ |
| | Dropout | 0.1 | 0.1, 0.2, 0.3 | 0.1 |

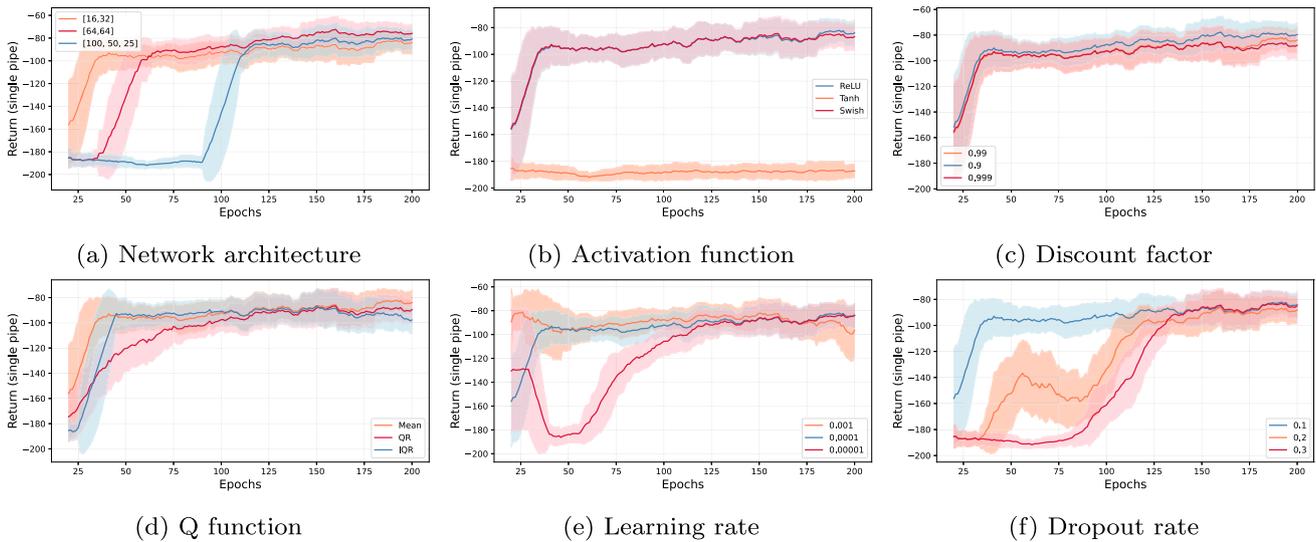

(a) Network architecture     (b) Activation function     (c) Discount factor

(d) Q function     (e) Learning rate     (f) Dropout rate

**Fig. 4** Significance of various parameters on the performance of offline agent: Each point is a rolling mean, and the shaded area is a rolling standard deviation return with the sliding window of 20 episodes

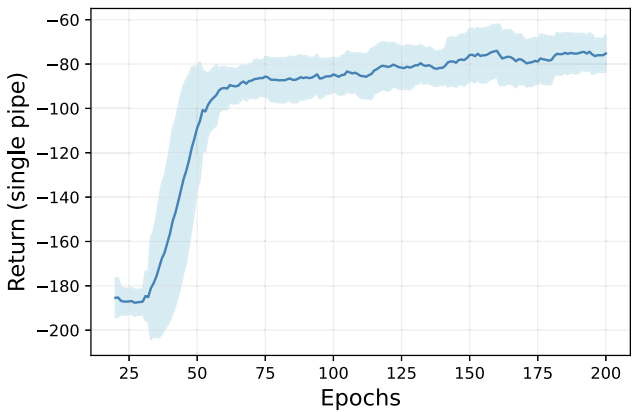

**Fig. 5** The training performance of offline CQL agent based on DQN replay buffer. The graph reports the mean and standard deviation of return with the sliding window of 20 epochs.

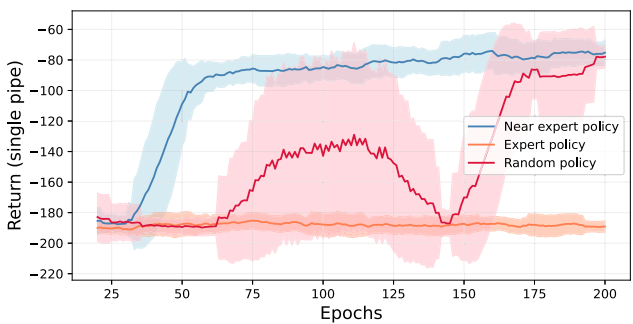

**Fig. 6** The learning performance of offline CQL agent with different source datasets collected using a random, a near-expert, and the expert policy

Without any intervention, the failure probability of assets will reach its limit, i.e., 1. We also compare the unit cost per method needed to reduce the failure probability in Fig. 8. The result shows that the CQL method performs slightly better in reducing the failure probability, followed by the DQN method. The better performance of CQL can be the result of using a near-expert policy obtained by the DQN agent. Both DRL-based methods outperform the preventive, corrective, and greedy maintenance approaches. The comparisons warrant the effectiveness of maintenance planning of the water pipes network using the DRL framework.

## 5.4 Discussion and limitations

We propose a novel DRL-based solution for the rehabilitation planning of water pipes. The proposed approach accommodates multiple pipes with distinct characteristics such as length, material, and failure rates. We show that the DRL-based methods, both online and offline agents, develop an optimal rehabilitation plan that ensures sufficient reduced failure probability with minimum cost. Depending on the requirements of the infrastructure managers, the reward function can be modified to ensure that none of the asset experience failure probability above a defined threshold.

In this paper, we, for the first time, investigate the offline reinforcement learning paradigm to solve the rehabilitation

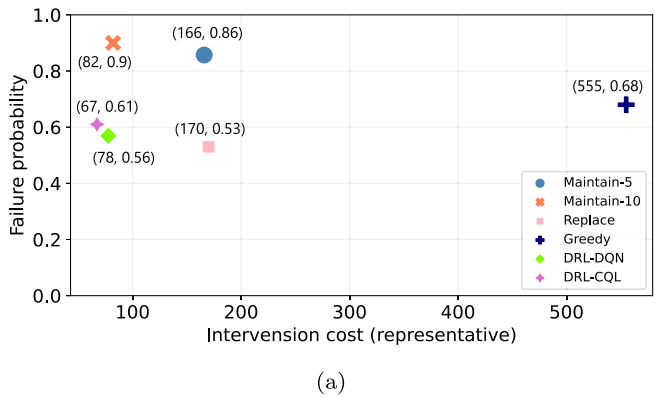

(a)

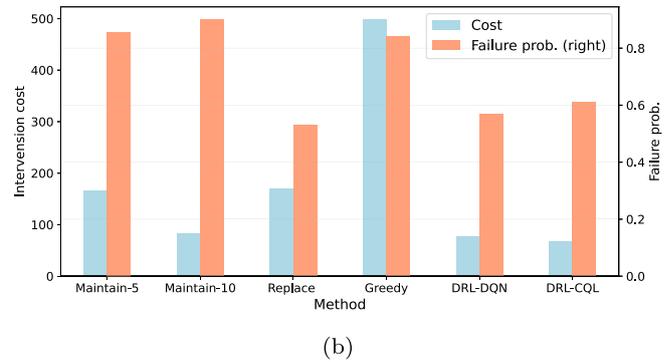

(b)

**Fig. 7** Comparison of online and offline DRL agents with baselines. We compare an average intervention cost and failure probability obtained for all 16 pipes using the trained DRL agents and baseline

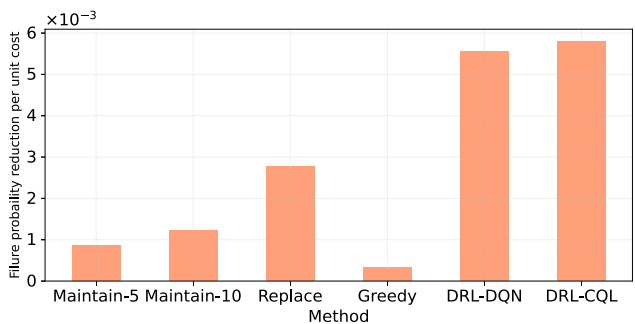

**Fig. 8** Comparison of unit cost for failure probability reduction. The DRL-based method results in the highest reduction of failure probability per unit cost

planning problem. Standard pipe maintenance datasets are an ideal candidate to evaluate the capabilities of offline reinforcement learning. This is because the typical maintenance dataset can be framed as a Markov decision process, were given a state (condition) of an asset, an action (intervention) is applied. Based on the chosen action, the state of an asset is either improved or remains the same. This setup requires the reward function, which can be formulated based on the improvement in the condition state of an asset. To illustrate offline reinforcement learning, we used the dataset generated by online interactions with the simulated environment based on near-expert DQN policy. The future work aims to use the larger dataset of water pipes based on different policies to explore the capabilities of offline reinforcement learning compared to online DRL.

Our case study uses a limited number of attributes related to each pipe: age, material, length, failure rate, and failure probability. However, several additional attributes can be added depending on the requirements of the decision-maker. The setup can be extended to include network zones, the number of previous failures, traffic load, the impact of unavailability, uncertainty arising from sudden policies. The lower cost and failure probability is preferred. The DRL-based method outperforms other alternatives. Within DRL agents, the CQL agent performs slightly better than DQN failures, and so on. Additionally, our approach assumes the assets are independent of each other. The proposed rehabilitation plan can be further improved by accounting for the co-located assets where maintenance is performed for network segments instead of a single pipe.

# 6 Conclusion and future work

We present a successful application of deep reinforcement learning (DRL) with an online deep Q-network (DQN) and an offline conservative Q-learning (CQL) method for the management of the water pipes network. The DRL-based trained agents effectively outperform classical planning approaches, mainly preventive, corrective, and greedy strategies, without the need for explicit expert knowledge and detailed heuristic rules. Besides optimal policy, the DRL framework yields transparency in rehabilitation planning due to distinct definitions of states, actions, and reward functions.

The future work of this study will extend this solution to include a comprehensive deterioration model, intervention costs, the impact of the interventions, and failure on the availability and surrounding households. The goal would be to devise a rehabilitation policy for underground utilities, including water and sewer pipes, to cluster intervention moments. The DRL is a promising framework for modeling sequential decision-making problems. However, it remains an under-explored research area in asset infrastructure management due to the complexity of modeling environments for multi-component structures. The offline DRL approach provides a favorable solution for such problem settings.

**Acknowledgements** This research has been partially funded by NWO under the grant PrimaVera NWA.1 160.18.238 and by the ERC Starting Grant 101077178 (DEUCE).

**Author Contributions** All authors contributed equally to the study's conception and design. Material preparation, data collection, and analysis were performed by the first author. The draft of the manuscript was written by the first author and all authors reviewed and corrected multiple versions of the manuscript. All authors read and approved the final manuscript.

## Declarations

**Conflict of interest** The authors have no conflicts of interest to declare that are relevant to the content of this article.